\documentclass[letterpaper]{article} 
\usepackage{aaai2026}  
\usepackage{times}  
\usepackage{helvet}  
\usepackage{courier}  
\usepackage[hyphens]{url}  
\usepackage{graphicx} 
\usepackage{natbib}  
\usepackage{caption} 
\usepackage{algorithm}
\usepackage{algorithmic}

\usepackage{multicol, multirow, threeparttable}
\usepackage{amsmath}
\usepackage{amsthm}
\usepackage{amssymb}
\usepackage{booktabs}
\usepackage{pifont}
\usepackage{newfloat}
\usepackage{listings}
\DeclareCaptionStyle{ruled}{labelfont=normalfont,labelsep=colon,strut=off} 
\floatstyle{ruled}
\newfloat{listing}{tb}{lst}{}
\floatname{listing}{Listing}
\usepackage{subcaption}
\usepackage{xspace}
\usepackage{xcolor}         
\usepackage{marvosym}

\definecolor{bl}{rgb}{0.25, 0.5, 0.9}
\newcommand{\best}[1]{{\textbf{\textcolor{red}{#1}}}}
\newcommand{\second}[1]{{\textcolor{bl}{\underline{#1}}}}

\usepackage[capitalize]{cleveref}
\crefname{section}{Sec.}{Secs.}
\Crefname{section}{Section}{Sections}
\Crefname{table}{Table}{Tables}
\crefname{table}{Tab.}{Tabs.}

\title{SitLLM: Large Language Models for Sitting Posture Health Understanding via Pressure Sensor Data}
\author{
    Jian Gao\textsuperscript{\rm 1,},
    Fufangchen Zhao\textsuperscript{\rm 1,},
    Yiyang Zhang\textsuperscript{\rm 1,}, 
    Danfeng Yan\textsuperscript{\rm 1}
}
\affiliations{
    \textsuperscript{\rm 1}Beijing University of Posts and Telecommunications \\
    \{gaojian, zhaofufangchen, zhangyiyang2025, yandf\}@bupt.edu.cn,
    
}

\usepackage{bibentry}

\begin{document}

\maketitle

\begin{abstract}
Poor sitting posture is a critical yet often overlooked factor contributing to long-term musculoskeletal disorders and physiological dysfunctions. Existing sitting posture monitoring systems, although leveraging visual, IMU, or pressure-based modalities, often suffer from coarse-grained recognition and lack the semantic expressiveness necessary for personalized feedback. In this paper, we propose \textbf{SitLLM}, a lightweight multimodal framework that integrates flexible pressure sensing with large language models (LLMs) to enable fine-grained posture understanding and personalized health-oriented response generation. SitLLM comprises three key components: (1) a \textit{Gaussian-Robust Sensor Embedding Module} that partitions pressure maps into spatial patches and injects local noise perturbations for robust feature extraction; (2) a \textit{Prompt-Driven Cross-Modal Alignment Module} that reprograms sensor embeddings into the LLM’s semantic space via multi-head cross-attention using the pre-trained vocabulary embeddings; and (3) a \textit{Multi-Context Prompt Module} that fuses feature-level, structure-level, statistical-level, and semantic-level contextual information to guide instruction comprehension. To enable high-quality cross-modal alignment, we constructed a pressure–text alignment dataset for sitting posture health, comprising 9,255 samples. The dataset was built through a two-stage process combining expert annotation, medical knowledge retrieval, and human–LLM collaborative generation. Extensive experiments demonstrate that SitLLM outperforms existing baselines on multiple sitting health-related tasks, including sitting posture analysis, sitting health-aware question answering, and sitting posture correction suggestions, while maintaining efficiency suitable for edge deployment. Our work highlights the potential of combining low-dimensional sensor modalities with semantically rich LLMs for personalized health monitoring.
\end{abstract}

\section{Introduction}

Prolonged poor sitting posture has been widely associated with a range of chronic health issues, including cervical spondylosis, lumbar disc herniation, musculoskeletal pain, and compromised cardiopulmonary function \cite{robertson2009effects}. Importantly, these health risks often originate not from overall posture but from long-term asymmetry or abnormal pressure on specific body regions such as the neck, lower back, or pelvis. For instance, forward head posture (FHP) alters cervical spine biomechanics by increasing cantilever loads and soft tissue strain, which has been associated with upper cervical joint degeneration and various clinical symptoms including cervicogenic pain and dizziness.\cite{chu2020plausible}. Thus, timely detection of localized postural anomalies and targeted feedback is essential for preventing long-term health deterioration and improving the granularity of posture-based health management.

\begin{figure}[!t]
    \centering
    \includegraphics[width=1.0\linewidth]{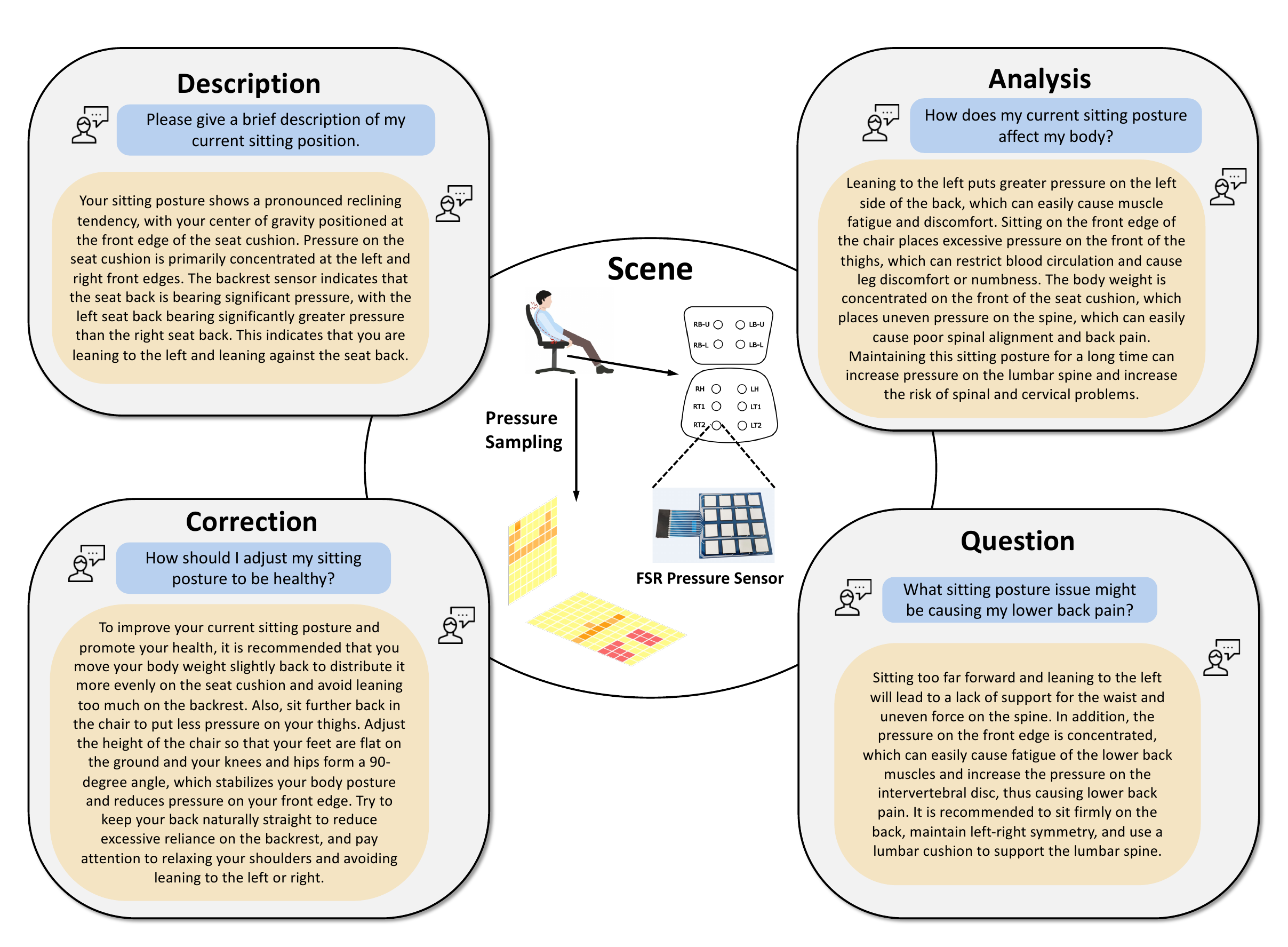}
    \caption{SitLLM understands pressure sensor inputs to respond to various posture-related health tasks.}
    \label{fig:ablation-pca}
\end{figure}

With increasing awareness of posture-related health concerns, researchers have developed various intelligent sitting posture monitoring systems using visual sensing\cite{fang2022sprnet, zhao2024sitting}, inertial measurement units (IMUs)\cite{tang2021upper}, and flexible pressure sensors\cite{yoon2024effective}. These systems aim to promote healthy behavior in sedentary contexts. However, most existing approaches rely on coarse-grained classification strategies (e.g., "standard posture," "leaning forward," "reclining") \cite{yongxiang2019sitting, cai2021improved} and simplistic interaction mechanisms such as fixed prompts or audiovisual alerts\cite{fan2022deep, la2023iot}. Such designs fall short in real-world scenarios characterized by diverse user needs and complex, dynamically changing posture states.

Current sitting posture recognition systems face two fundamental limitations: (1) insufficient perceptual granularity, which hampers the detection of subtle anomalies in key anatomical regions such as the shoulders, lower back, and pelvis; and (2) lack of semantic feedback, as most systems rely on templated, low-level responses that cannot support personalized recommendations or multi-layered health-oriented dialogues.

Among various sensing modalities, flexible pressure sensors have emerged as a compelling solution for continuous and stable posture monitoring, owing to their non-invasive nature and ease of integration. However, pressure distribution alone lacks the semantic expressiveness necessary for high-level understanding and personalized interaction.

To address the limitations in semantic expressiveness, we introduce a large language model to interpret pressure sensor data and generate personalized responses.Despite recent advances in LLMs' multi-task semantic understanding and generation capabilities, applying LLMs directly to posture modeling remains challenging due to three core factors:

\begin{itemize}
\item Modality Gap and Semantic Misalignment: Raw pressure maps are inherently low-dimensional and unstructured, lacking explicit semantic annotations\citep{liang2022mind, deldari2022cocoa}. This makes it difficult for LLMs to establish meaningful spatial-semantic correspondences, often leading to representation errors and semantic drift.

\item Limited Alignment Capability \& High Deployment Cost: Mainstream LLMs are typically large in scale and computationally expensive, posing difficulties for real-time deployment on resource-limited devices. Meanwhile, smaller models, while efficient, suffer from reduced expressiveness and weakened cross-modal alignment, severely limiting downstream performance\citep{xu2024device}.

\item Domain Knowledge Deficiency: General-purpose LLMs lack specialized knowledge in posture-related health risks, which impedes their ability to identify abnormal postures and generate actionable, clinically relevant feedback.
\end{itemize}

To address these challenges, we propose SitLLM, a lightweight multimodal framework that integrates flexible pressure sensing with LLM-based semantic modeling to support fine-grained sitting posture understanding and personalized feedback generation. Specifically, our contributions include:

\begin{itemize}
\item \textbf{Gaussian-Robust Sensor Embedding Module:} We adopt a ViT-style block-wise encoding strategy with Gaussian perturbation to enhance spatial robustness.

\item \textbf{Sensor Queries Alignment Module:} This module bridges the modality gap by projecting sensor embeddings as queries into the LLM’s vocabulary space through multi-head cross-attention, achieving deep semantic alignment without requiring large-scale paired pretraining.

\item \textbf{Multi-Context Prompt Module:} We design a hierarchical prompt mechanism that injects feature-level, structure-level, statistical-level, and instruction-level sensor semantics into the LLM. This allows the model to integrate multimodal cues and improve response generation across diverse sitting health tasks.

\item \textbf{Instruction-Tuned Dataset with Medical Knowledge Injection:} We construct a fine-grained sensor–text dataset that covers four types of posture-related tasks and 120 typical sitting postures. Our two-stage generation pipeline incorporates expert annotations, retrieval-augmented medical knowledge, and GPT-4o-based semantic generation.
\end{itemize}

Experimental results demonstrate that our proposed framework significantly outperforms existing methods in representative tasks such as posture description, personalized suggestion generation, and posture-related health Q\&A. It achieves superior accuracy, generalization, and semantic interaction capabilities. These findings validate the effectiveness of integrating flexible pressure sensing with lightweight LLMs under resource constraints, and demonstrate the potential of our approach to support real-time, multi-task, health-aware intelligent agents.

\section{Related Work}

\subsection{Aligning Sensor with Language}
Although establishing semantic connections between sensor data and natural language has emerged as a growing research focus, constructing large-scale, high-quality, and semantically aligned cross-modal corpora from unstructured sensor signals remains a significant challenge. Such corpora are essential for training high-performance cross-modal models \cite{li2024sensorllm,chen2024sensor2text,imran2024llasa}.

Existing studies generally adopt two main strategies to address this issue. The first involves generating textual summaries of sensor features using fixed template rules or incorporating raw sensor values as structured prompts into downstream LLM-based tasks \cite{kim2024health}. The second strategy employs dedicated sensor encoders combined with alignment modules to enhance the model’s capacity for interpreting and reasoning over sensor-derived information \cite{imran2024llasa, li2024sensorllm}.

However, these approaches are fundamentally constrained by the sparsity and limited semantic coverage of available sensor-text paired data, which impedes both the semantic alignment capability and the generalization performance of cross-modal models across diverse tasks.

\begin{figure*}[!t]
\centering
\includegraphics[width=1\textwidth]{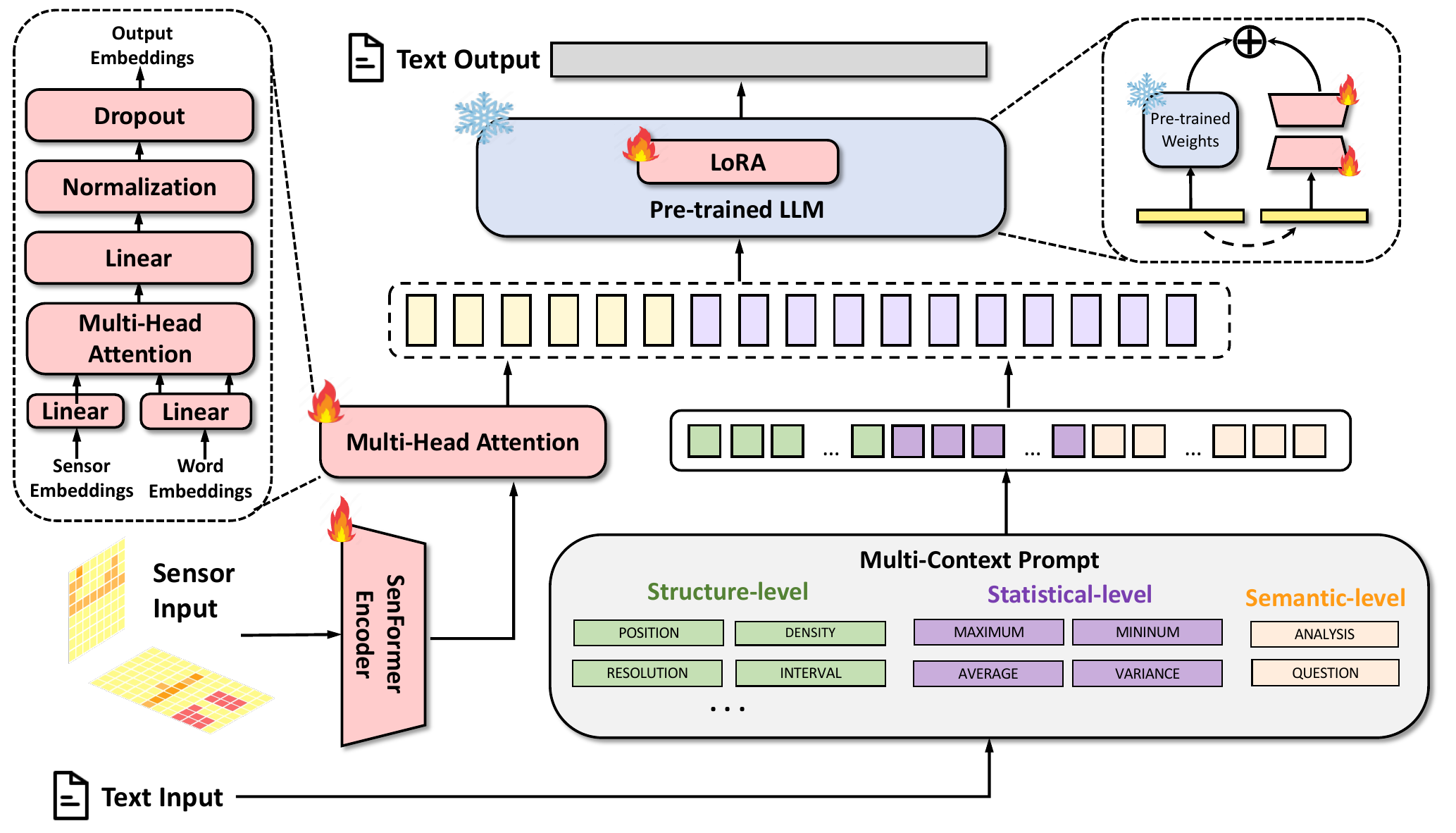}
\caption{
Overview of the SitLLM architecture. The framework integrates three core components: (1) the \textbf{Sensor Embedding Module}, which transforms pressure maps into spatially-aware embeddings; (2) the \textbf{Sensor Queries Alignment Module}, which reprograms sensor embeddings into the LLM’s vocabulary space through multi-head cross-attention; and (3) the \textbf{Multi-Context Prompt Module}, which injects multi-level semantic contexts into the language model.}
\label{fig:pipeline}
\end{figure*}

\subsection{Health LLM}
In recent years, large language models (LLMs) have demonstrated remarkable capabilities in medical language understanding and reasoning, driving advances in tasks such as intelligent consultation and health-related question answering\citep{singhal2023large,nori2023capabilities}. Studies on models like PaLM 2 and GPT-4 have shown that domain-specific fine-tuning and instruction tuning can endow LLMs with the ability to outperform traditional methods in various medical benchmarks, exhibiting causal reasoning, personalized generation, and clinical inference.

With the rise of multimodal learning, researchers have extended LLMs to medical imaging modalities (e.g., X-rays, MRIs), resulting in visual-language models such as Med-Flamingo\citep{moor2023med} and LLaVA-Med\citep{li2023llava}. However, these efforts predominantly focus on static images, while the modeling of temporal physiological signals and wearable sensor data remains in its nascent stage.

Recent work has begun to explore the application of LLMs to low-dimensional physiological modalities such as ECG and PPG \citep{feli2025llm}, incorporating prompt engineering and structured prompting mechanisms to enhance semantic alignment. Concurrently, models like Health-LLM\citep{kim2024health}, LLaSA\citep{imran2024llasa}, and Sensor2Text\citep{chen2024sensor2text} have expanded the cross-modal capabilities of LLMs in domains such as activity recognition, mental health assessment, and multi-turn health dialogues. These efforts mark a shift from image-centric modeling toward sensor-informed semantic understanding.

Nevertheless, most of these approaches still rely on limited sensor–text paired data and lack a unified modeling framework for encoding structural priors, task types, and semantic preferences. This significantly hinders their generalization and interpretability in complex health scenarios involving dynamic behavior, personalized feedback, and semantic reasoning.

\section{Model Structure}

SitLLM is composed of three key modules:(1) Sensor Embedding Module, which encodes pressure modality data into compact and structure-aware feature vectors;(2) Sensor Queries Alignment Module, which aligns sensor embeddings with the textual semantic space, enabling deep semantic connectivity with the LLM;(3) Multi-Context Prompt Module, which injects sensor semantic information into the backbone language model through multi-level contextual prompting to generate the final response.An overview of the overall architecture is shown in Figure~\ref{fig:pipeline}.

\subsection{Sensor Embedding Module}

We adopt a Vision Transformer (ViT)-style encoding strategy to process the input pressure map. Given a two-dimensional pressure matrix $P \in \mathbb{R}^{H \times W}$, we first divide it into fixed-size patches $p_i \in \mathbb{R}^{s \times s}$. To enhance the model’s robustness against sensor noise and subtle posture perturbations, we introduce a Gaussian perturbation mechanism during training, adding random noise to each patch:

$$
p_i' = p_i + \mathcal{N}(0, \sigma^2 I)
$$

where $\sigma$ is a tunable hyperparameter controlling the intensity of the Gaussian noise. Each perturbed patch is then flattened and passed through a linear projection to obtain its embedding representation $e_i \in \mathbb{R}^d$. To retain spatial positional information, a learnable positional encoding $E_{\text{pos}} \in \mathbb{R}^{N \times d}$ is added. The final embedding sequence is then obtained by:

$$
X = \text{Transformer}(\text{Concat}(e_i + E_{\text{pos}})) \in \mathbb{R}^{N \times d}
$$

where $N = \frac{H \times W}{s^2}$ denotes the number of patches. The Transformer encoder is used to model both local and long-range spatial dependencies.

\subsection{Sensor Queries Alignment Module}

Due to the lack of available aligned pre-trained models for sensor modalities, directly mapping sensor representations into the language space remains a significant challenge. We propose an unsupervised reprogramming mechanism that treats the sensor embeddings $X \in \mathbb{R}^{N \times d}$ as queries and projects them into the LLM's pretrained vocabulary embedding space $E \in \mathbb{R}^{V \times d}$, where $V$ denotes the vocabulary size. Cross-modal semantic alignment is achieved through a multi-head cross-attention mechanism. For the $k$-th attention head, the projections are defined as:

$$
Q_k = X W_k^Q, \quad K_k = E W_k^K, \quad V_k = E W_k^V
$$

where $W_k^Q, W_k^K, W_k^V \in \mathbb{R}^{d \times d}$ are learnable projection matrices. The attention output for head $k$ is computed as:

$$
Z_k = \text{Attention}(Q_k, K_k, V_k) = \text{Softmax}\left(\frac{Q_k K_k^\top}{\sqrt{d}}\right)V_k
$$

The outputs of all heads are concatenated to obtain the fused representation:

$$
Z = \text{Concat}(Z_1, Z_2, ..., Z_K) \in \mathbb{R}^{N \times d}
$$

Finally, $Z$ is projected through a linear transformation, followed by normalization and dropout, to match the hidden dimension $D$ of the LLM backbone, yielding the aligned features $O \in \mathbb{R}^{N \times D}$.

\subsection{Multi-context Prompt Module}

Prompting is the most direct approach to injecting modality-specific knowledge into large language models (LLMs). In multimodal LLMs, it is common to wrap the encoded image features with pad tokens and insert them as a prefix before the textual input to guide generation\citep{zhu2023minigpt, li2023blip}. However, for sensor modalities, the absence of large-scale aligned training data leads to incomplete semantic representations, making simple prefix concatenation insufficient to effectively activate the reasoning capability of LLMs.

To address this issue, we design a multi-context prompt mechanism that incorporates multi-level semantic cues from the sensor modality, including:

Feature-level context: Preserves the raw semantic features of the sensor modality by directly using the aligned embeddings as token-level prompts, guiding the model to attend to localized perceptual information.

Structure-level context: Encodes spatial layout and dynamic characteristics of the sensor array, such as sensor positions, distribution density, sampling intervals, and resolution.

Statistical-level context: Extracts global statistical summaries (e.g., maximum pressure, minimum pressure, mean, variance), enabling the model to build a high-level understanding of the input.

Semantic-level context primarily encodes the user’s task instructions, explicitly guiding the language model to comprehend the current task intent. It works in conjunction with the aforementioned feature-level, structure-level, and statistical-level sensor semantics to generate responses that align with the instruction semantics.

These contextual cues are fused with natural language instructions through prompt engineering, forming a composite input that guides the backbone LLM to effectively leverage supplemental semantic information from the sensor modality during instruction interpretation. Compared to conventional single-level prompting, the multi-context prompt structure offers more efficient and comprehensive representation, effectively mitigating performance bottlenecks caused by insufficient cross-modal alignment.

\section{Sensor-Language Dataset Construction}

\begin{figure}[!t]
    \centering
    \includegraphics[width=1\linewidth]{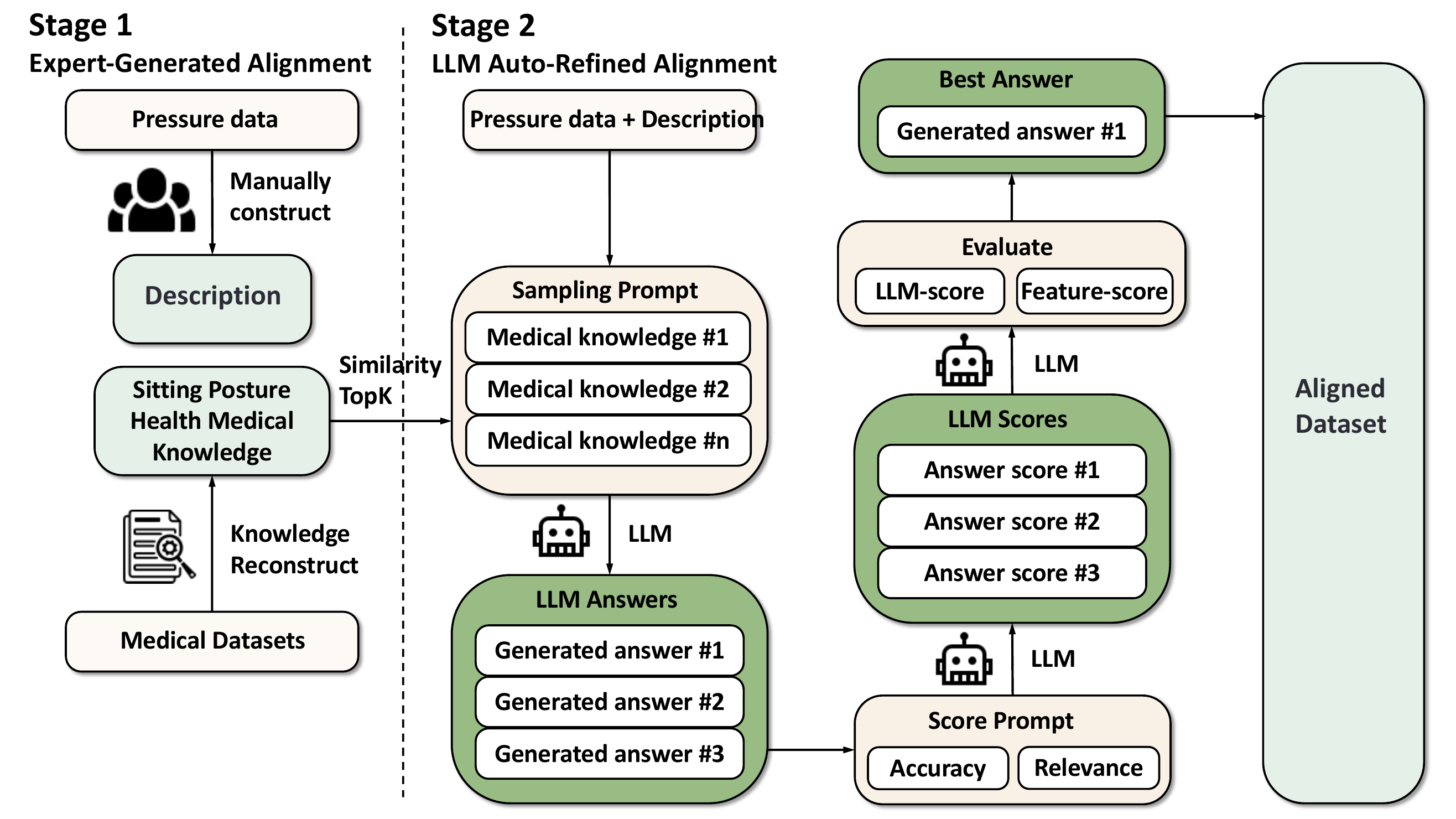}
    \caption{
    Overview of our two-stage sensor–language dataset construction pipeline.}
    \label{fig:data-create}
\end{figure}

Existing datasets often suffer from coarse-grained classification\citep{clever2020bodies, yuan2021smart, gelaw2021posture}, lacking fine-grained aligned textual descriptions corresponding to pressure sensor data, and missing professional medical knowledge related to sitting posture health. These limitations restrict the model’s ability to perform fine-grained understanding of sitting posture features represented by pressure data, hinder its generalization across various sitting health instruction tasks, and limit its professionalism in addressing medical health-related question answering tasks.

To address these issues, we propose a sensor–text alignment construction method that incorporates professional medical health knowledge. Based on this method, we introduce medical knowledge datasets such as HuaTuo-26M\citep{li2023huatuo} to build a high-quality sitting pressure–text alignment dataset containing 9,255 samples, covering four task types, and roughly categorized into 120 posture types. An overview of this process is illustrated in Figure~\ref{fig:data-create}.

The dataset includes not only fine-grained posture descriptions (e.g., the support condition of the backrest on the upper left back, the degree of contact between the right thigh and the seat surface), but also various types of sitting health tasks, such as posture problem analysis, posture correction suggestions, and sitting health question answering.

Our proposed construction method consists of two stages. 

\subsection{Stage 1}
First, we recruited eight volunteers with diverse heights and body types to collect a total of 1,063 valid pressure samples under various sitting postures. During data collection, three experienced experts simultaneously provided fine-grained descriptive annotations for each posture, producing high-quality textual representations. Based on variations across different body regions, these samples can be categorized into approximately 120 typical sitting postures, forming a solid foundation for subsequent task-aligned data construction.

Second, we built a structured knowledge base covering disease types potentially caused by improper sitting postures. Based on this knowledge base, we screened relevant instruction-tuning (SFT) samples from medical datasets such as HuaTuo-26M\citep{li2023huatuo}, DISC-Med-SFT\cite{bao2023discmedllm}, and FreedomIntelligence/HuatuoGPT2-SFT-GPT4-140K\citep{chen2023huatuogptii}. Using the all-MiniLM-L6-v2\citep{wang2020minilm} model, we encoded the instruction texts into semantic vectors to construct a retrievable sitting posture health knowledge base, which was later used to inject professional medical knowledge into our dataset.

\subsection{Stage 2}

In the second stage, we employed GPT-4o to automatically generate pressure–text alignment samples. Specifically, we first randomly sampled an expert-annotated pressure instance and its corresponding posture description as the input example, prompting the model to generate a task-specific question (e.g., posture analysis, health recommendation). We then encoded this question into a semantic vector and retrieved the Top10 most relevant knowledge entries from the medical knowledge base based on cosine similarity. Among these, five entries were randomly selected as domain knowledge prompts. These, along with the original posture description, were embedded into the prompt to guide GPT-4o in generating three candidate responses.

To evaluate response quality, we designed a scoring prompt that guided the model to assign scores from two aspects: accuracy and relevance, yielding an LLM-based score $S_{\mathrm{llm}}$. Additionally, we employed keyword extraction to identify sitting posture features in the responses and compared them against those in the original description, computing both keyword coverage and semantic similarity to obtain a feature score $S_{\mathrm{feat}}$. This metric reflects whether irrelevant or incorrect features were introduced.

Finally, we combined the two scores using a weighted sum to compute the final score:

$$
S_{\mathrm{final}} = \alpha \cdot S_{\mathrm{llm}} + (1 - \alpha) \cdot S_{\mathrm{feat}},
$$

and selected the highest-scoring response to be included in the final sitting pressure–text alignment dataset.

\subsection{Parameter Efficient Training}

To mitigate catastrophic forgetting and improve training efficiency, we adopt a parameter-efficient fine-tuning strategy to optimize the pre-trained large language model (LLM). Specifically, we incorporate Low-Rank Adaptation (LoRA)\citep{hu2022lora} into the backbone model to enable lightweight tuning of key modules, while applying full-parameter fine-tuning to other trainable components in a joint optimization manner. During training, the primary objective is the supervised loss $L_{\text{sup}}$, computed between the model's output and the reference answer.

\section{Experiments}


\renewcommand{\arraystretch}{1.0}
\begin{table*}[htb]
\setlength{\tabcolsep}{3pt}
\scriptsize
\centering
\begin{threeparttable}
\begin{tabular}{c|cccc|cccc|cccc|cccc}
\toprule
\multirow{2}{*}{\textbf{Model}} & 
\multicolumn{4}{c|}{\textbf{Analysis}} & 
\multicolumn{4}{c|}{\textbf{Correction}} & 
\multicolumn{4}{c|}{\textbf{Description}} & 
\multicolumn{4}{c}{\textbf{Question}} \\
\cmidrule(lr){2-5} \cmidrule(lr){6-9} \cmidrule(lr){10-13} \cmidrule(lr){14-17}
 & BLEU & R-L & BERT & METEOR & BLEU & R-L & BERT & METEOR & BLEU & R-L & BERT & METEOR & BLEU & R-L & BERT & METEOR \\
\midrule
Qwen2.5-1.5B-Instruct & 0.0157 & 0.1971 & 0.7084 & 0.1346 & 0.0170 & 0.1900 & 0.6989 & 0.1422 & 0.0165 & 0.1735 & 0.5966 & 0.1347 & 0.0596 & 0.1891 & 0.7188 & 0.2594 \\
Qwen2.5-72B-Instruct & 0.0388 & 0.2027 & 0.7435 & 0.2099 & 0.0476 & 0.1922 & 0.7428 & 0.2493 & 0.0463 & 0.1934 & 0.7447 & 0.2325 & \second{0.0680} & \second{0.2276} & \second{0.7382} & \second{0.3103} \\
Qwen-Max & 0.0397 & 0.1826 & 0.7389 & 0.2220 & \second{0.0477} & 0.1827 & \second{0.7430} & \second{0.2567} & \best{0.0532} & 0.1928 & \second{0.7481} & \second{0.2628} & 0.0502 & 0.2070 & 0.7241 & 0.2952 \\
DeepSeek-V3 & 0.0301 & \second{0.2240} & 0.7376 & \second{0.2431} & 0.0212 & 0.1696 & 0.7169 & 0.1940 & 0.0240 & 0.1880 & 0.7230 & 0.1997 & 0.0364 & 0.2085 & 0.7097 & 0.2587 \\
DeepSeek-R1 & 0.0308 & 0.2135 & 0.7301 & 0.2079 & 0.0275 & 0.1798 & 0.7183 & 0.1954 & 0.0321 & 0.2057 & 0.7280 & 0.2156 & 0.0622 & 0.2269 & 0.7240 & 0.2745 \\
GPT-4o & \second{0.0446} & \best{0.2283} & 0.7441 & 0.2082 & 0.0457 & \second{0.1967} & 0.7382 & 0.2297 & 0.0472 & \best{0.2086} & 0.7431 & 0.2272 & 0.0548 & 0.2109 & 0.7305 & 0.3052 \\
Gemini-2.5-Pro & 0.0302 & 0.2173 & \second{0.7492} & 0.2311 & 0.0315 & 0.1750 & 0.7190 & 0.2087 & 0.0428 & 0.1893 & 0.7329 & 0.2458 & 0.0509 & 0.2144 & 0.7341 & 0.2952 \\
\midrule
SitLLM-1.5B & \best{0.0698} & 0.2185 & \best{0.7563} & \best{0.3010} & 
\best{0.0527} & \best{0.2138} & \best{0.7662} & \best{0.3317} & \second{0.0507} & \second{0.2081} & \best{0.7505} & \best{0.3061} & \best{0.0891} & \best{0.2473} & \best{0.7602} & \best{0.3154} \\
\bottomrule
\end{tabular}

\caption{Cross-model evaluation on pressure-sensor Q\&A tasks. Results for Analysis, Correction, Description, and Question tasks are reported across four metrics: BLEU, ROUGE-L (R-L), BERTScore (BERT), and METEOR.}
\label{tab::text-results}
\end{threeparttable}
\end{table*}

\paragraph{\textbf{Dataset.}}We evaluate the model’s capabilities across multiple core tasks in the sitting posture health domain, including fine-grained posture perception, posture risk analysis, correction suggestion generation, and health-related question answering. To this end, we construct an evaluation dataset comprising 1,000 samples, collected from two individuals who did not participate in the training process, ensuring objectivity and generalization of the testing phase. The construction of the evaluation data strictly follows the same filtering criteria and generation pipeline as the pre-training dataset, covering four representative task categories with 250 samples per task.

\paragraph{Implementation Details.} We adopt Qwen2.5-1.5B-Instruct\citep{qwen2.5} as the backbone large language model and perform parameter-efficient fine-tuning using the LoRA technique, with the rank parameter set to $r=64$ and the scaling factor $\text{lora}_\alpha = 16$. The optimization is conducted using the Adam optimizer\citep{kingma2014adam} with an initial learning rate of $3\times10^{-5}$. The model is trained for 5 epochs with a batch size of 4, gradient accumulation steps of 8, and a maximum sequence length of 4096.

\subsection{Generation Quality Evaluation}

\paragraph{\textbf{Baselines.}} Currently, there is no unified evaluation benchmark for posture-related question answering (QA) tasks based on pressure sensor data. To address this gap, we designed an evaluation protocol that leverages the prompt engineering capabilities of large language models (LLMs) to construct QA tasks. Specifically, we inject pressure-sensing information into the model through structured prompts, including key attributes such as sensor placement, value range, sampling precision, and the spatial intervals between pressure points. Additionally, raw pressure data is encoded into textual form and embedded within the prompt to guide the model in understanding pressure distribution for task completion. We conduct benchmark testing across several representative LLMs, including the open-source models Qwen2.5-72B-Instruct, DeepSeek-V3 and DeepSeek-R1, the proprietary models GPT-4o, Qwen-Max, and Gemini 2.5 Pro, as well as the lightweight backbone model Qwen2.5-1.5B-Instruct, which serves as the foundation of our system.

\paragraph{\textbf{Metric.}} We employ a variety of mainstream natural language processing evaluation metrics to assess the quality of the model-generated outputs. Specifically, BLEU\citep{papineni2002bleu}, ROUGE-L\citep{lin2004rouge}, and METEOR\citep{banerjee2005meteor} are used as fundamental metrics to measure surface-level similarity and n-gram overlap between the generated text and reference answers. Additionally, to further evaluate semantic consistency, we incorporate BertScore\citep{zhang2019bertscore}, which leverages a pretrained BERT model to compute the cosine similarity between contextual embeddings of the generated and reference texts, thereby effectively capturing deeper semantic matching.

\paragraph{\textbf{Results.}} Detailed evaluation results for all models are presented in \Cref{tab::text-results}. Notably, compared to the strongest general-purpose LLM, GPT-4o, our SitLLM-1.5B achieves significant improvements of 56.5\% and 44.5\% in BLEU and METEOR scores, respectively, on the core \textbf{Analysis} task. When compared to a lightweight model of the same scale, Qwen2.5-1.5B-Instruct, we observe an overall performance gain of over 83.7\% across all tasks. Furthermore, in both the \textbf{Correction} and \textbf{Question} tasks, our method surpasses all baseline models, achieving key metric advantages ranging from 10.5\% to 18.6\%.

\subsection{Semantic Quality Evaluation}

\paragraph{\textbf{Baselines.}} We perform semantic quality evaluation using the same baseline models and prompt-based generation methodology as adopted in the Generation Quality Evaluation section.

\renewcommand{\arraystretch}{0.8}
\begin{table}[!t]
    \setlength{\tabcolsep}{4pt}
    \centering
    \scriptsize
    \begin{threeparttable}
    \begin{tabular}{c|cc|cc|cc|cc}
    \toprule
         \multicolumn{1}{c}{\multirow{2}{*}{\scalebox{1.2}{Models}}} & \multicolumn{2}{c}{Analysis} & \multicolumn{2}{c}{Correction} & \multicolumn{2}{c}{Description} & \multicolumn{2}{c}{Question} \\

          \cmidrule(lr){2-3} \cmidrule(lr){4-5} \cmidrule(lr){6-7} \cmidrule(lr){8-9}
         
         \multicolumn{1}{c}{} & G-S & H-S & G-S & H-S & G-S & H-S & G-S & H-S \\
    \toprule

   Qwen2.5-1.5B & 0.548 & 0.682 & 0.572 & 0.704 & 0.418 & 0.626 & 0.606 & 0.686 \\
    
    Qwen2.5-72B & 0.756 & 0.762 & 0.768 & 0.744 & 0.740 & 0.744 & 0.738 & 0.756 \\
    
    Qwen-Max & 0.740 & 0.754 & 0.776 & 0.786 & 0.674 & 0.706 & 0.734 & 0.776 \\

    DeepSeek-V3 & 0.702 & 0.696 & 0.748 & 0.766 & 0.708 & 0.762 & 0.714 & 0.702 \\

    DeepSeek-R1   & 0.754 & \second{0.782} & 0.776 & 0.726 & 0.740 & \best{0.786} & 0.760 & 0.728 \\

    GPT-4o & \second{0.784} & 0.754 & \second{0.786} & \second{0.804} & \second{0.746} & 0.754 & 0.728 & 0.748 \\

    Gemini-2.5-Pro   & 0.762 & 0.744 & 0.754 & 0.742 & 0.722 & 0.754 & \second{0.762} & \second{0.794} \\

    SitLLM-1.5B & \best{0.788} & \best{0.806} & \best{0.802} & \best{0.822} & \best{0.764} & \second{0.782} & \best{0.798} & \best{0.834} \\

    \bottomrule
    
    \end{tabular}
    \caption{Semantic alignment performance across four sitting posture health tasks, evaluated using GPT-Score (G-S) and expert Human-Score (H-S).}

    \label{tab:semantic}
    \end{threeparttable}
\end{table}
\paragraph{\textbf{Metric.}} We adopt both automated LLM-based metrics (GPT-Score) and expert human evaluations (Human-Score), guided by a unified, domain-specific rubric.

GPT-Score utilizes GPT-4o as an expert evaluator via a customized prompt tailored to the sitting posture health domain. The model assesses the alignment between generated and reference answers by analyzing key domain-relevant concepts such as affected body regions, pressure distribution patterns, posture-related health risks, and correction strategies. Scores are assigned on a continuous scale from 0 to 1, based on semantic accuracy, content coverage, and the presence of critical domain-specific terminology. This follows the established "LLM-as-a-judge" paradigm \citep{zheng2023judging}.
Human-Score is produced by posture health experts using the same rubric as the GPT-based evaluator. Experts rate responses on a 0–1 scale, ensuring consistency in evaluation criteria while incorporating professional domain knowledge.

\paragraph{\textbf{Results.}} Table~\ref{tab:semantic} shows that SitLLM-1.5B achieves the best semantic alignment across all tasks, outperforming strong baselines including GPT-4o and Qwen2.5-72B by up to 3.8\% in GPT-Score and 4.0\% in Human-Score. Gains are especially notable in Correction and Question tasks, evidencing superior domain understanding and response accuracy. High consistency between GPT-Score and Human-Score validates the effectiveness of our automated evaluation.These results demonstrate that our domain-specific fine-tuning and prompt design effectively enhance the model’s semantic understanding and alignment with expert knowledge, thus enabling more accurate and clinically meaningful posture health assessments.

\renewcommand{\arraystretch}{1.0}
\begin{table*}[!t]
    \setlength{\tabcolsep}{4pt}
    \centering
    \scriptsize
    \begin{threeparttable}
    \begin{tabular}{p{\textwidth}}
    \toprule
    \textbf{Question} \\
    \textit{I feel a lot of pressure in the front of my thighs when I sit. Will this affect blood circulation in my legs and possibly cause varicose veins or numbness in my lower limbs? How can I prevent these problems?} \\
    \midrule
    \textbf{GPT-4o Answer} \\
    The increased pressure on the front of the thighs can indeed affect blood circulation in the legs. Maintaining this posture for long periods of time can lead to varicose veins or leg numbness. To prevent these problems, it's recommended to adjust your sitting posture to distribute pressure and avoid excessive pressure on a single area. You can also use a more ergonomic chair and increase regular movement and stretching exercises to improve circulation. Regularly standing up and moving around, changing your posture every hour, and avoiding sitting in the same position for extended periods can help reduce leg discomfort and prevent related problems. \\
    \midrule
    \textbf{SitLLM-1.5B Answer} \\
    Yes, excessive pressure on the front of the thighs may affect blood circulation in the legs, increasing the risk of varicose veins or leg numbness. \textcolor{red}{You put too much weight on the front edge of the seat cushion, and your legs are under great pressure.} This sitting posture compresses the blood vessels in the thigh area, resulting in poor blood return. Over time, it may cause varicose veins in the lower limbs or leg numbness. To prevent these problems, you first need to adjust your sitting posture. \textcolor{red}{Make sure your entire buttocks are sitting deep on the chair, avoid putting your center of gravity on the front edge of the seat cushion.} At the same time, \textcolor{red}{keep your back in a normal curvature and avoid leaning back too much.} You can consider \textcolor{red}{adjusting the height of the chair and the tilt angle of the seat cushion to keep the knee and hip joints at about 90 degrees.} In addition, get up and move around every 30 minutes to promote blood circulation in the lower limbs. If your work requires long periods of sitting, you can stretch your leg muscles regularly or gently shake your ankles. \\
    \bottomrule
    \end{tabular}
    \caption{Case study comparing GPT-4o and SitLLM in answering a posture-related health question. SitLLM provides more personalized and structured recommendations, with key points marked in \textcolor{red}{red}.}
    \label{tab:case-study-sitllm}
    \end{threeparttable}
\end{table*}

\renewcommand{\arraystretch}{0.8}
\begin{table}[t]
    \setlength{\tabcolsep}{4pt}
    \scriptsize
    \centering
    \begin{tabular}{lccc}
    \toprule
    \textbf{Metrics} & \textbf{Qwen2.5-1.5B-Instruct} & \textbf{Qwen2.5-72B-Instruct} & \textbf{SitLLM-1.5B} \\
    \midrule
    BLEU & 0.0596 & \second{0.0680} & \best{0.0705} \\
    ROUGE-L & 0.1891 & \best{0.2276} & \second{0.2098} \\
    BERTScore & 0.7188 & \best{0.7382} & \second{0.7221} \\
    METEOR & 0.2594 & \best{0.3103} & \second{0.2761} \\
    GPT-Score & 0.6060 & \second{0.7380} & \best{0.7980} \\
    Human-Score & 0.6860 & \second{0.7560} & \best{0.8340} \\
    \bottomrule
    \end{tabular}
    \caption{Zero-shot learning performance comparison on the \textit{question} task. Results show SitLLM's cross-task transfer capability when trained only on \textit{description}, \textit{analysis}, and \textit{correction} tasks.}
    \label{tab:zero-shot}
\end{table}

\subsection{Zero-shot Learning}

\paragraph{\textbf{Setup.}}Large language models typically exhibit strong zero-shot task generalization capabilities. To evaluate whether SitLLM demonstrates similar performance on unseen sitting posture health tasks, we designed a cross-task generalization experiment. Specifically, SitLLM is trained only on data from three task types: \textit{description}, \textit{analysis}, and \textit{correction}, using the same training configurations and hyperparameters as the full-data setting. Without accessing any training data from the \textit{question} task, we evaluate SitLLM on its corresponding test set to assess its zero-shot generalization performance.

\paragraph{\textbf{Baselines.}}
To comprehensively evaluate the zero-shot task generalization capability of SitLLM, we conduct a comparative analysis against two baseline models: Qwen2.5-1.5B-Instruct, which is of a similar scale, and Qwen2.5-72B-Instruct, which achieves the best performance on the \textit{question} task.

\paragraph{\textbf{Metric.}}
To comprehensively evaluate the zero-shot generalization ability of SitLLM, we adopt both the standard natural language generation metrics used in the Generation Quality Evaluation (BLEU, ROUGE-L, BERTScore, and METEOR) and the semantic quality metrics from the Semantic Quality Evaluation (GPT-Score and Human-Score). 

\paragraph{\textbf{Results.}} As shown in Table~\ref{tab:zero-shot}, SitLLM-1.5B demonstrates strong zero-shot generalization ability on the unseen \textbf{question} task, even without access to any question-specific training data. Compared to Qwen2.5-1.5B-Instruct, SitLLM achieves superior performance across all six evaluation metrics, with notable improvements in both generation quality and semantic quality. While Qwen2.5-72B-Instruct benefits from significantly larger model capacity, SitLLM-1.5B still outperforms it on key generation metrics such as BLEU and GPT-Score.

\subsection{Case Study}
To qualitatively assess the effectiveness of SitLLM, we conducted a comparative analysis against GPT-4o in the context of real-world sitting posture health scenarios (Table~\ref{tab:case-study-sitllm}). While GPT-4o tends to provide general health suggestions, SitLLM demonstrates a finer-grained understanding of user posture by explicitly referencing pressure patterns sensed from the seat (e.g., excessive pressure on the front edge of the cushion). Based on this sensory insight, SitLLM generates targeted correction suggestions, such as adjusting hip positioning, modifying lumbar support, or shifting weight distribution and seat angle. These results highlight SitLLM's capacity to deliver more fine-grained, personalized, and expert-level responses by leveraging multimodal pressure data for sitting posture reasoning and health guidance.

\section{Conclusion}
We presented SitLLM, a lightweight multimodal framework that integrates pressure sensing with large language models for sitting posture understanding and health feedback. By introducing robust sensor embedding, prompt-based semantic alignment, and structured instruction injection, SitLLM achieves accurate, interpretable, and task-adaptive posture analysis. To support training and evaluation, we constructed a high-quality multimodal dataset with diverse pressure–text pairs. Extensive experiments across four tasks confirm the effectiveness of SitLLM under low-resource settings and its superiority over strong baselines. Future work will explore the integration of diverse sensor modalities and their application to broader health-related scenarios beyond posture, including intelligent medical monitoring and assistive interventions.

\clearpage

\bibliography{SitLLM}

\begin{thebibliography}{39}
\providecommand{\natexlab}[1]{#1}

\bibitem[{Banerjee and Lavie(2005)}]{banerjee2005meteor}
Banerjee, S.; and Lavie, A. 2005.
\newblock METEOR: An automatic metric for MT evaluation with improved correlation with human judgments.
\newblock In \emph{Proceedings of the acl workshop on intrinsic and extrinsic evaluation measures for machine translation and/or summarization}, 65--72.

\bibitem[{Bao et~al.(2023)Bao, Chen, Xiao, Ren, Wu, Zhong, Peng, Huang, and Wei}]{bao2023discmedllm}
Bao, Z.; Chen, W.; Xiao, S.; Ren, K.; Wu, J.; Zhong, C.; Peng, J.; Huang, X.; and Wei, Z. 2023.
\newblock DISC-MedLLM: Bridging General Large Language Models and Real-World Medical Consultation.
\newblock arXiv:2308.14346.

\bibitem[{Cai et~al.(2021)Cai, Zhao, Zhang, Xu, and Li}]{cai2021improved}
Cai, W.; Zhao, D.; Zhang, M.; Xu, Y.; and Li, Z. 2021.
\newblock Improved self-organizing map-based unsupervised learning algorithm for sitting posture recognition system.
\newblock \emph{Sensors}, 21(18): 6246.

\bibitem[{Chen et~al.(2023)Chen, Wang, Gao, Jiang, Chen, Zhang, Song, Xie, Kong, Li, Wan, Li, and Wang}]{chen2023huatuogptii}
Chen, J.; Wang, X.; Gao, A.; Jiang, F.; Chen, S.; Zhang, H.; Song, D.; Xie, W.; Kong, C.; Li, J.; Wan, X.; Li, H.; and Wang, B. 2023.
\newblock HuatuoGPT-II, One-stage Training for Medical Adaption of LLMs.
\newblock arXiv:2311.09774.

\bibitem[{Chen et~al.(2024)Chen, Cheng, Wang, Zhao, and Matusik}]{chen2024sensor2text}
Chen, W.; Cheng, J.; Wang, L.; Zhao, W.; and Matusik, W. 2024.
\newblock Sensor2text: Enabling natural language interactions for daily activity tracking using wearable sensors.
\newblock \emph{Proceedings of the ACM on Interactive, Mobile, Wearable and Ubiquitous Technologies}, 8(4): 1--26.

\bibitem[{Chu, Lo, and Bhaumik(2020)}]{chu2020plausible}
Chu, E.~C.; Lo, F.~S.; and Bhaumik, A. 2020.
\newblock Plausible impact of forward head posture on upper cervical spine stability.
\newblock \emph{Journal of family medicine and primary care}, 9(5): 2517--2520.

\bibitem[{Clever et~al.(2020)Clever, Erickson, Kapusta, Turk, Liu, and Kemp}]{clever2020bodies}
Clever, H.~M.; Erickson, Z.; Kapusta, A.; Turk, G.; Liu, K.; and Kemp, C.~C. 2020.
\newblock Bodies at rest: 3d human pose and shape estimation from a pressure image using synthetic data.
\newblock In \emph{Proceedings of the IEEE/CVF conference on computer vision and pattern recognition}, 6215--6224.

\bibitem[{Deldari et~al.(2022)Deldari, Xue, Saeed, Smith, and Salim}]{deldari2022cocoa}
Deldari, S.; Xue, H.; Saeed, A.; Smith, D.~V.; and Salim, F.~D. 2022.
\newblock Cocoa: Cross modality contrastive learning for sensor data.
\newblock \emph{Proceedings of the ACM on Interactive, Mobile, Wearable and Ubiquitous Technologies}, 6(3): 1--28.

\bibitem[{Fan et~al.(2022)Fan, Hu, Chen, Zhang, and Ma}]{fan2022deep}
Fan, Z.; Hu, X.; Chen, W.-M.; Zhang, D.-W.; and Ma, X. 2022.
\newblock A deep learning based 2-dimensional hip pressure signals analysis method for sitting posture recognition.
\newblock \emph{Biomedical Signal Processing and Control}, 73: 103432.

\bibitem[{Fang et~al.(2022)Fang, Shi, Fang, and Yin}]{fang2022sprnet}
Fang, Y.; Shi, S.; Fang, J.; and Yin, W. 2022.
\newblock SPRNet: sitting posture recognition using improved vision transformer.
\newblock In \emph{2022 International Joint Conference on Neural Networks (IJCNN)}, 1--6. IEEE.

\bibitem[{Feli et~al.(2025)Feli, Azimi, Liljeberg, and Rahmani}]{feli2025llm}
Feli, M.; Azimi, I.; Liljeberg, P.; and Rahmani, A.~M. 2025.
\newblock An LLM-Powered Agent for Physiological Data Analysis: A Case Study on PPG-based Heart Rate Estimation.
\newblock \emph{arXiv preprint arXiv:2502.12836}.

\bibitem[{Gelaw and Hagos(2021)}]{gelaw2021posture}
Gelaw, T.~A.; and Hagos, M.~T. 2021.
\newblock Posture prediction for healthy sitting using a smart chair.
\newblock In \emph{International Conference on Advances of Science and Technology}, 401--411. Springer.

\bibitem[{Hu et~al.(2022)Hu, Shen, Wallis, Allen-Zhu, Li, Wang, Wang, Chen et~al.}]{hu2022lora}
Hu, E.~J.; Shen, Y.; Wallis, P.; Allen-Zhu, Z.; Li, Y.; Wang, S.; Wang, L.; Chen, W.; et~al. 2022.
\newblock Lora: Low-rank adaptation of large language models.
\newblock \emph{ICLR}, 1(2): 3.

\bibitem[{Imran et~al.(2024)Imran, Khan, Biswas, and Islam}]{imran2024llasa}
Imran, S.~A.; Khan, M. N.~H.; Biswas, S.; and Islam, B. 2024.
\newblock Llasa: Large multimodal agent for human activity analysis through wearable sensors.
\newblock \emph{arXiv preprint arXiv:2406.14498}, 3(4).

\bibitem[{Kim et~al.(2024)Kim, Xu, McDuff, Breazeal, and Park}]{kim2024health}
Kim, Y.; Xu, X.; McDuff, D.; Breazeal, C.; and Park, H.~W. 2024.
\newblock Health-llm: Large language models for health prediction via wearable sensor data.
\newblock \emph{arXiv preprint arXiv:2401.06866}.

\bibitem[{Kingma and Ba(2014)}]{kingma2014adam}
Kingma, D.~P.; and Ba, J. 2014.
\newblock Adam: A method for stochastic optimization.
\newblock \emph{arXiv preprint arXiv:1412.6980}.

\bibitem[{La~Mura et~al.(2023)La~Mura, De~Gregorio, Lamberti, and Tucci}]{la2023iot}
La~Mura, M.; De~Gregorio, M.; Lamberti, P.; and Tucci, V. 2023.
\newblock Iot system for real-time posture asymmetry detection.
\newblock \emph{Sensors}, 23(10): 4830.

\bibitem[{Li et~al.(2023{\natexlab{a}})Li, Wong, Zhang, Usuyama, Liu, Yang, Naumann, Poon, and Gao}]{li2023llava}
Li, C.; Wong, C.; Zhang, S.; Usuyama, N.; Liu, H.; Yang, J.; Naumann, T.; Poon, H.; and Gao, J. 2023{\natexlab{a}}.
\newblock Llava-med: Training a large language-and-vision assistant for biomedicine in one day.
\newblock \emph{Advances in Neural Information Processing Systems}, 36: 28541--28564.

\bibitem[{Li et~al.(2023{\natexlab{b}})Li, Li, Savarese, and Hoi}]{li2023blip}
Li, J.; Li, D.; Savarese, S.; and Hoi, S. 2023{\natexlab{b}}.
\newblock Blip-2: Bootstrapping language-image pre-training with frozen image encoders and large language models.
\newblock In \emph{International conference on machine learning}, 19730--19742. PMLR.

\bibitem[{Li et~al.(2023{\natexlab{c}})Li, Wang, Wu, Zhang, Xu, Fu, Tiwari, Wan, and Wang}]{li2023huatuo}
Li, J.; Wang, X.; Wu, X.; Zhang, Z.; Xu, X.; Fu, J.; Tiwari, P.; Wan, X.; and Wang, B. 2023{\natexlab{c}}.
\newblock Huatuo-26m, a large-scale chinese medical qa dataset.
\newblock \emph{arXiv preprint arXiv:2305.01526}.

\bibitem[{Li et~al.(2024)Li, Deldari, Chen, Xue, and Salim}]{li2024sensorllm}
Li, Z.; Deldari, S.; Chen, L.; Xue, H.; and Salim, F.~D. 2024.
\newblock Sensorllm: Aligning large language models with motion sensors for human activity recognition.

\bibitem[{Liang et~al.(2022)Liang, Zhang, Kwon, Yeung, and Zou}]{liang2022mind}
Liang, V.~W.; Zhang, Y.; Kwon, Y.; Yeung, S.; and Zou, J.~Y. 2022.
\newblock Mind the gap: Understanding the modality gap in multi-modal contrastive representation learning.
\newblock \emph{Advances in Neural Information Processing Systems}, 35: 17612--17625.

\bibitem[{Lin(2004)}]{lin2004rouge}
Lin, C.-Y. 2004.
\newblock Rouge: A package for automatic evaluation of summaries.
\newblock In \emph{Text summarization branches out}, 74--81.

\bibitem[{Moor et~al.(2023)Moor, Huang, Wu, Yasunaga, Dalmia, Leskovec, Zakka, Reis, and Rajpurkar}]{moor2023med}
Moor, M.; Huang, Q.; Wu, S.; Yasunaga, M.; Dalmia, Y.; Leskovec, J.; Zakka, C.; Reis, E.~P.; and Rajpurkar, P. 2023.
\newblock Med-flamingo: a multimodal medical few-shot learner.
\newblock In \emph{Machine Learning for Health (ML4H)}, 353--367. PMLR.

\bibitem[{Nori et~al.(2023)Nori, King, McKinney, Carignan, and Horvitz}]{nori2023capabilities}
Nori, H.; King, N.; McKinney, S.~M.; Carignan, D.; and Horvitz, E. 2023.
\newblock Capabilities of gpt-4 on medical challenge problems.
\newblock \emph{arXiv preprint arXiv:2303.13375}.

\bibitem[{Papineni et~al.(2002)Papineni, Roukos, Ward, and Zhu}]{papineni2002bleu}
Papineni, K.; Roukos, S.; Ward, T.; and Zhu, W.-J. 2002.
\newblock Bleu: a method for automatic evaluation of machine translation.
\newblock In \emph{Proceedings of the 40th annual meeting of the Association for Computational Linguistics}, 311--318.

\bibitem[{Robertson et~al.(2009)Robertson, Amick~III, DeRango, Rooney, Bazzani, Harrist, and Moore}]{robertson2009effects}
Robertson, M.; Amick~III, B.~C.; DeRango, K.; Rooney, T.; Bazzani, L.; Harrist, R.; and Moore, A. 2009.
\newblock The effects of an office ergonomics training and chair intervention on worker knowledge, behavior and musculoskeletal risk.
\newblock \emph{Applied ergonomics}, 40(1): 124--135.

\bibitem[{Singhal et~al.(2023)Singhal, Azizi, Tu, Mahdavi, Wei, Chung, Scales, Tanwani, Cole-Lewis, Pfohl et~al.}]{singhal2023large}
Singhal, K.; Azizi, S.; Tu, T.; Mahdavi, S.~S.; Wei, J.; Chung, H.~W.; Scales, N.; Tanwani, A.; Cole-Lewis, H.; Pfohl, S.; et~al. 2023.
\newblock Large language models encode clinical knowledge.
\newblock \emph{Nature}, 620(7972): 172--180.

\bibitem[{Tang et~al.(2021)Tang, Tan, Su, Chiang, and Chen}]{tang2021upper}
Tang, H.-Y.; Tan, S.-H.; Su, T.-Y.; Chiang, C.-J.; and Chen, H.-H. 2021.
\newblock Upper body posture recognition using inertial sensors and recurrent neural networks.
\newblock \emph{Applied Sciences}, 11(24): 12101.

\bibitem[{Team(2024)}]{qwen2.5}
Team, Q. 2024.
\newblock Qwen2.5: A Party of Foundation Models.

\bibitem[{Wang et~al.(2020)Wang, Wei, Dong, Bao, Yang, and Zhou}]{wang2020minilm}
Wang, W.; Wei, F.; Dong, L.; Bao, H.; Yang, N.; and Zhou, M. 2020.
\newblock Minilm: Deep self-attention distillation for task-agnostic compression of pre-trained transformers.
\newblock \emph{Advances in neural information processing systems}, 33: 5776--5788.

\bibitem[{Xu et~al.(2024)Xu, Li, Chen, Wang, Gao, Cai, and Ling}]{xu2024device}
Xu, J.; Li, Z.; Chen, W.; Wang, Q.; Gao, X.; Cai, Q.; and Ling, Z. 2024.
\newblock On-device language models: A comprehensive review.
\newblock \emph{arXiv preprint arXiv:2409.00088}.

\bibitem[{Yongxiang et~al.(2019)Yongxiang, Jingle, Sanpeng, Yuming, Peng, Zijing, and Tianjiang}]{yongxiang2019sitting}
Yongxiang, J.; Jingle, D.; Sanpeng, D.; Yuming, Q.; Peng, W.; Zijing, W.; and Tianjiang, Z. 2019.
\newblock Sitting posture recognition by body pressure distribution and airbag regulation strategy based on seat comfort evaluation.
\newblock \emph{The Journal of Engineering}, 2019(23): 8910--8914.

\bibitem[{Yoon, Moon, and Son(2024)}]{yoon2024effective}
Yoon, J.; Moon, A.; and Son, S.~W. 2024.
\newblock Effective Posture Classification Using Statistically Significant Data from Flexible Pressure Sensors.
\newblock \emph{IEEE Journal on Flexible Electronics}, 3(5): 173--180.

\bibitem[{Yuan and Li(2021)}]{yuan2021smart}
Yuan, L.; and Li, J. 2021.
\newblock Smart cushion based on pressure sensor array for human sitting posture recognition.
\newblock In \emph{2021 IEEE Sensors}, 1--4. IEEE.

\bibitem[{Zhang et~al.(2019)Zhang, Kishore, Wu, Weinberger, and Artzi}]{zhang2019bertscore}
Zhang, T.; Kishore, V.; Wu, F.; Weinberger, K.~Q.; and Artzi, Y. 2019.
\newblock Bertscore: Evaluating text generation with bert.
\newblock \emph{arXiv preprint arXiv:1904.09675}.

\bibitem[{Zhao and Su(2024)}]{zhao2024sitting}
Zhao, S.; and Su, Y. 2024.
\newblock Sitting Posture Recognition Based on the Computer's Camera.
\newblock In \emph{Proceedings of the 2024 2nd Asia Conference on Computer Vision, Image Processing and Pattern Recognition}, 1--5.

\bibitem[{Zheng et~al.(2023)Zheng, Chiang, Sheng, Zhuang, Wu, Zhuang, Lin, Li, Li, Xing et~al.}]{zheng2023judging}
Zheng, L.; Chiang, W.-L.; Sheng, Y.; Zhuang, S.; Wu, Z.; Zhuang, Y.; Lin, Z.; Li, Z.; Li, D.; Xing, E.; et~al. 2023.
\newblock Judging llm-as-a-judge with mt-bench and chatbot arena.
\newblock \emph{Advances in neural information processing systems}, 36: 46595--46623.

\bibitem[{Zhu et~al.(2023)Zhu, Chen, Shen, Li, and Elhoseiny}]{zhu2023minigpt}
Zhu, D.; Chen, J.; Shen, X.; Li, X.; and Elhoseiny, M. 2023.
\newblock Minigpt-4: Enhancing vision-language understanding with advanced large language models.
\newblock \emph{arXiv preprint arXiv:2304.10592}.

\end{thebibliography}

\end{document}